# Comprehensive Evaluation of Machine Learning for Type 2 Diabetes Risk Prediction: Large-Scale External Validation and Fairness Analysis

Rajveer Singh Pall[1], Sameer Yadav*[2], Siddharth Bhalerao[2], Sourabh Sahu[3], Ritu Ahluwalia[1], Bhaskar Awadhiya[4]
[1]Department of Computer Science & Business Systems, Gyan Ganga Institute of Technology and Sciences, Jabalpur, India
[2]Department of Artificial Intelligence & Machine Learning, Gyan Ganga Institute of Technology and Sciences, Jabalpur, India
[3]Department of Artificial Intelligence & Robotics, Gyan Ganga Institute of Technology and Sciences, Jabalpur, India
[4]Department of Electronics & Communication, Manipal Institute of Technology, MAHE, Manipal, 576104, India

***Abstract*—Machine learning-based Type 2 diabetes risk prediction models mostly obtain good internal validation results but lose their effectiveness in real-world applications due to deficient external testing and fairness assessment. We developed a comprehensive, multi-dimensional framework for evaluating discrimination, calibration, interpretability, and algorithmic fairness when applied to nationally representative populations. An XGBoost model and baseline models were trained on NHANES 2015-2020 (n = 15,685) using eight non-laboratory predictors: age, sex, race/ethnicity, BMI, smoking status, physical activity, history of heart attack, and history of stroke. External validation was performed with BRFSS 2020–2022 (n=1,285,783), a phone-based surveillance system that has realistic deployment conditions and a large distribution shift. Internal validation showed good discrimination (AUC 0.794, 95% CI 0.788 to 0.800), but there was some loss of performance on external validation (AUC=0.717, relative decrease: −9.7%, $p < 0.001$). Fairness analysis revealed severe bias: predictions for elderly adults (≥60) were much worse than for young adults (AUC 0.607 vs 0.742, difference 0.135, $p < 0.001$), the obese underperformed compared to normal weight individuals (AUC 0.698 vs 0.735, difference of 0.037). Gender demonstrated comparability (male 0.723 versus female 0.712, $p = 0.142$). Calibration showed that there was an overestimation of the risk (Brier score 0.123). The SHAP analysis showed that age, BMI, and physical activity were the main risk drivers with a clinically consistent direction. Internal validation severely overestimates real-life performance. Massive external validation revealed systematic errors in the most vulnerable: those who are old and obese, who have the highest risk of diabetes, get a worse AI-based score. Such differences call for the development of equity consideration-based and age-specific deployment approaches before any clinical use.**

***Index Terms*—Type 2 diabetes, machine learning, external validation, algorithmic fairness, XGBoost, health equity**

## I. Introduction

Today, T2DM affects 537 million adults worldwide, and this number is projected to reach 783 million by 2045; unfortunately, almost half of the cases are undiagnosed [1]. Early detection allows lifestyle interventions resulting in 58% reduc- tion in progression [2]. Machine learning offers the potential for better risk stratification, with published models achieving internal validation AUC greater than 0.85 [14]. However, three key gaps continue to hamper clinical implementation.

There are three reasons for this: firstly, many studies use small homogeneous datasets, which do not generalise. For example, the much-cited Pima Indians data set includes just 768 people from a single racial group. Second is the relative paucity of external validation—less than 15% of diabetes models are exposed to independent assessment in systematic reviews. Being trained on internal metrics, such models generally suffer from severe performance drop due to distribution shift across different populations, time periods or data collection procedures [3]. Third, evaluation is limited to discrimination measures such as AUC and ignores calibration performance, clinical utility at decision thresholds, or fairness of the algorithm across demographics [4]. This creates dangerous blind spots: a model may have a high overall AUC but systematically perform poorly for vulnerable populations, such as older adults or underrepresented racial minorities.

We overcome these limitations through four main innovations. First, we adopt a three-tier validation scheme: using NHANES (n=15,685) for development and BRFSS (n=1,285,783) for large-scale external testing under the realistic distribution shift from different sampling schemes and measurement protocols, as well as population characteristics. Second, we perform extensive fairness analysis over demographic and clinical subgroups exhibiting performance differences that are obfuscated by aggregated metrics. Third, we practise dimension-aware evaluation across discrimination, calibration, interpretability with SHAP values and fairness auditing. Fourth, we report in adherence with TRIPOD-AI [7] transparent reporting checklists for reproducibility.

Our emphasis on non-laboratory predictors serves the pragmatic reality that blood testing is inaccessible or unaffordable, allowing potential application in community health settings, pharmacies and telemedicine solutions.

*Corresponding author: Sameer Yadav (sameeryadav@ggits.org)

## II. METHODS

### A. Data Sources and Study Design

The development cohort was obtained from the National Health and Nutrition Examination Survey (NHANES) 2015-2020, which is a cross-sectional nationally representative data set via stratified multistage probability sampling with standardised physical examination and laboratory testing [8]. We limited analyses to participants aged ≥18 years who had complete information on predictors and the outcome (n=15,685). The validation cohort used exists in BRFSS 2020-2022, which is a telephone-based health survey with random-digit-dialling and cellular phone sampling [9]. This sample (n=1,285,783) brings real-world distributional shift from self-reported measures and a different demographic configuration, as well as a telephone-based assessment.

Diabetes was determined in NHANES as laboratory-measured HbA1c ≥6.5% or fasting plasma glucose ≥126 mg/dL or self-reported physician-diagnosed diabetes or use of current antidiabetic medication [6]. BRFSS diabetes status was
based on self-reported physician diagnosis. This measurement variation in development and validation is indicative of the real-world deployment scenario.

### B. Feature Selection

We limited the variables to eight non-biomarkers that were common in both datasets: age (years, continuous), gender (male/female), race/ethnicity (Non-Hispanic White, Non-Hispanic Black, Hispanic, Other), Body Mass Index (kg/m²; continuous), smoking status (never, former, current), physical activity (<150 minutes per week moderate-intensity activity or ≥ 150 min/week of moderate intensity exercise; binary), heart attack history (binary) and stroke history (binary). This non-laboratory-based testing makes it applicable to screening models that do not have laboratory facilities available, for instance, community settings, pharmacies and mobile health applications.

### C. Data Preprocessing

Median and mode were used to impute missing values in continuous and categorical variables, respectively. Winsorization of extreme outliers beyond three standard deviations was performed. Categorical variables underwent one-hot encoding. We normalised continuous features as z-scores to avoid scale-dependent bias. Class unequivalences were catered for by applying class weighting that is inversely proportional to the sample distribution in the corresponding classes.

### D. Model Development

We use XGBoost because it performed well with structured/tabular data and has built-in support for missing value imputation [10]. The comparison models were Logistic Regression (strong interpretable baseline), Random Forest (ensemble method), and a non-linear Support Vector Machine with RBF kernel as a strong non-linear baseline. We adopted a strict 5-fold nested cross-validation analysis on NHANES to avoid optimistic bias. Outer loop (left) for unbiased performance estimates to test folds and inner loop (right) for Bayesian hyperparameter optimisation, 100 iterations per fold.

The hyperparameter search space for XGBoost was defined as learning_rate from 0.01 to 0.20, max_depth from 3 to 9, n_estimators from 100 to 500, subsample from 0.6 to 1.0, colsample_bytree from 0.6 to 1.0 and min_child_weight in the range of [1–10]. The best hyperparameters: learning_rate=0.11, max_depth=6, n_estimators=240, subsample=0.85, colsample_bytree=0.80 and min_child_weight=3.

### E. Evaluation Framework

The discrimination metrics were the AUC-ROC with DeLong method 95% confidence intervals, sensitivity, specificity, positive predictive value, negative predictive value and F1- score. Youden's J statistic was used to identify the best oper- ating threshold. Calibration evaluation was based on the Brier score, which measured the mean-squared difference between predicted and observed values of probability, supplemented with calibration curves showing agreement between predicted risk (deciles of prognosis) and observed frequency. SHAP values broke down predictions into feature contributions, thus facilitating global ranking of important features and individual prediction explanations [11].

Fairness assessment was performed by gender (male, female), age category 18-39, 40-59, and ≥60 years, and BMI category (normal 18.5-24.9; overweight 25-29.9 kg/m²; obese ≥30 kg/m²). P-values of receiver operating characteristic curves were based on compared DeLong tests with Bonferroni correction. Fairness measures were disparate impact and equalised odds differences.

## III. RESULTS

### A. Cohort Characteristics and Distribution Shift

Table I presents baseline characteristics in development and validation cohorts, with a large distributional shift observed between the two groups. BRFSS participants were substantially older (mean 55.1 vs 49.0 years, $p < 0.001$) and had lower BMI (28.5 vs 29.7 kg/m², $p < 0.001$). The proportion of physical activity was very different ($p < 0.001$, 75.8% vs. 25.4%), probably due to methodological differences in the measurement. The prevalence of diabetes was less in BRFSS (13.3% versus 18.5%, $p < 0.001$), which could be attributed to self-reported bias or sampling variation.

### B. Internal Validation Performance

Table II shows the internal validation results with nested 5-fold cross-validation on NHANES. The tuned XGBoost produced the best discrimination (AUC 0.794, 95% CI 0.788-0.800), demonstrating superiority over all baseline models. XGBoost showed the sensitivity of 76.2%, specificity of 69.5% at the cutoff point determined with Youden's J statistic, and PPV of 43.8%, NPV of 90.1%, F1-score was 0.518, respec- tively. The high NPV (90.1%) suggests an excellent potential to exclude diabetes among low-risk individuals, making it suitable for screening uses.

TABLE I
BASELINE CHARACTERISTICS OF STUDY COHORTS

| Characteristic | NHANES (n=15,685) | BRFSS (n=1,285,783) | p-value |
|---|---|---|---|
| Age, years | 49.0 ± 18.6 | 55.1 ± 17.7 | $< 0.001$ |
| BMI, kg/m² | 29.7 ± 7.4 | 28.5 ± 6.5 | $< 0.001$ |
| Male, n (%) | 7,605 (48.5) | 596,991 (46.4) | $< 0.001$ |
| Female, n (%) | 8,080 (51.5) | 688,792 (53.6) | — |
| Ever Smoker, n (%) | 6,311 (40.2) | 487,117 (37.9) | $< 0.001$ |
| Physically Active[†] | 3,981 (25.4) | 974,180 (75.8) | $< 0.001$ |
| Diabetes, n (%) | 2,908 (18.5) | 170,868 (13.3) | $< 0.001$ |

[†]Physically active: ≥150 min/week moderate-intensity activity

TABLE II
INTERNAL VALIDATION PERFORMANCE ON NHANES

| Model | AUC (95% CI) | Sens. | Spec. | PPV | NPV |
|---|---|---|---|---|---|
| XGBoost | 0.794 (0.788–0.800) | 0.762 | 0.695 | 0.438 | 0.901 |
| Random Forest | 0.789 (0.783–0.795) | 0.755 | 0.688 | 0.431 | 0.898 |
| SVM (RBF) | 0.778 (0.772–0.784) | 0.748 | 0.672 | 0.415 | 0.893 |
| Logistic Reg. | 0.776 (0.770–0.782) | 0.741 | 0.669 | 0.409 | 0.891 |

Figure 1 displays ROC curves comparing all models on internal validation.

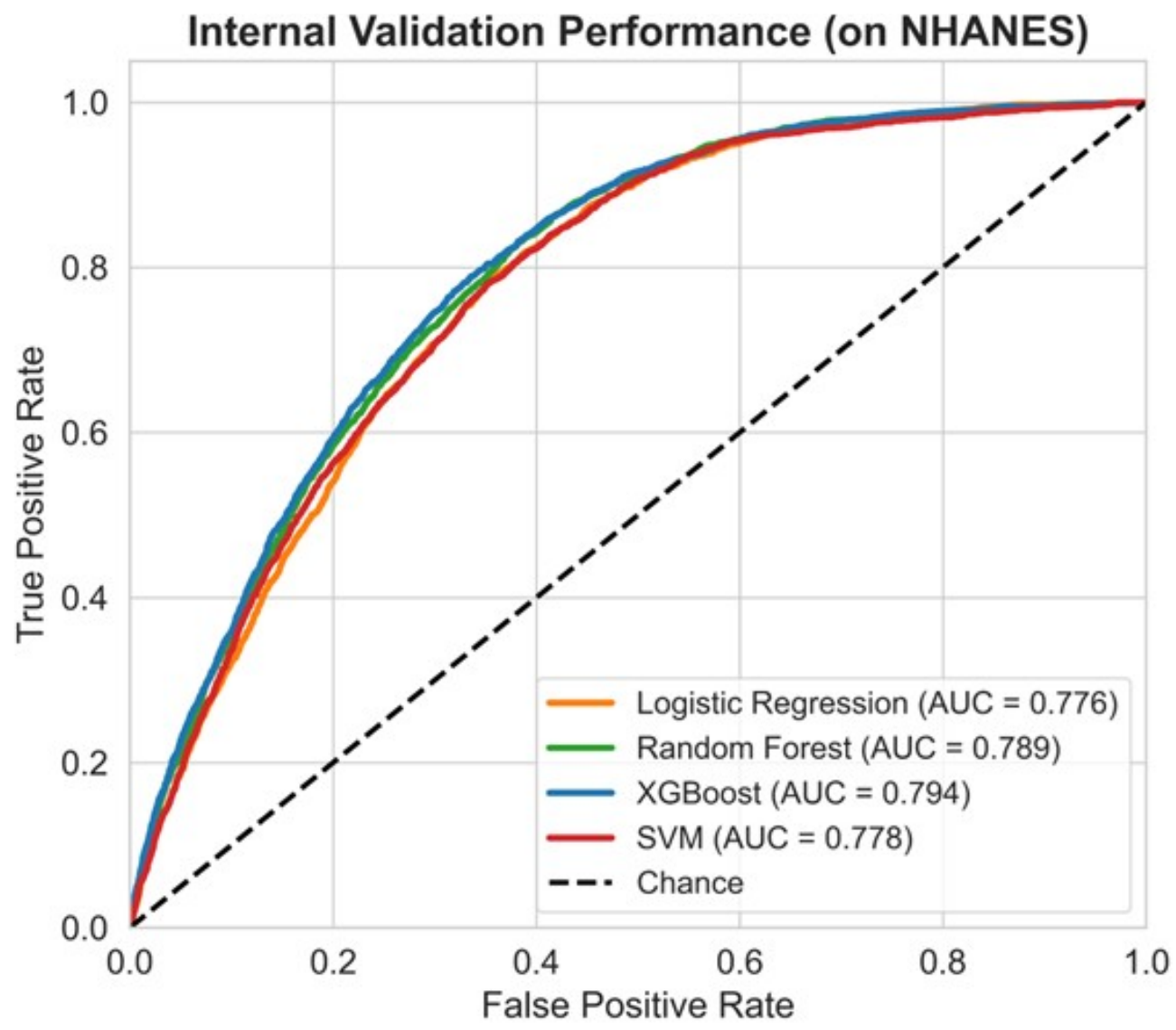

Fig. 1. ROC curves for internal validation on NHANES comparing Logistic Regression, Random Forest, XGBoost, and SVM.

### C. External Validation and Performance Degradation

XGBoost AUC decreases from 0.794 (internal) to 0.717 (external), which corresponds to an absolute drop of 0.077 (9.7% relative decline) relative difference ($p < 0.001$, DeLong test). The sensitivity, specificity, PPV and NPV of the external validation index were 68.7%, 67.2%, 22.3% and 94.1%. The F1-score on external validation was 0.336 at the external validation context.

The dissociation curve differential showed sensitivity decreasing more substantially (76.2% to 68.7%, $-7.5$ percentage points), whereas specificity decreased to a lesser extent (69.5% to 67.2%, $-2.3$ percentage points). The reason there is a 43.8% to 22.3% = 21.5% PPV drop is from a combination of lower disease prevalence (13.3% vs 18.5%), predictor distributions (distribution shift), and measurement noninvariance when comparing laboratory confirmation to self-report.

Figure 2 illustrates the reduction in external validation performance, showing that internal validation heavily overestimates real-world performance. The 9.7% relative decrease is in line with the reductions from systematic reviews showing median external AUC decreases of 10-12% for clinical prediction models.

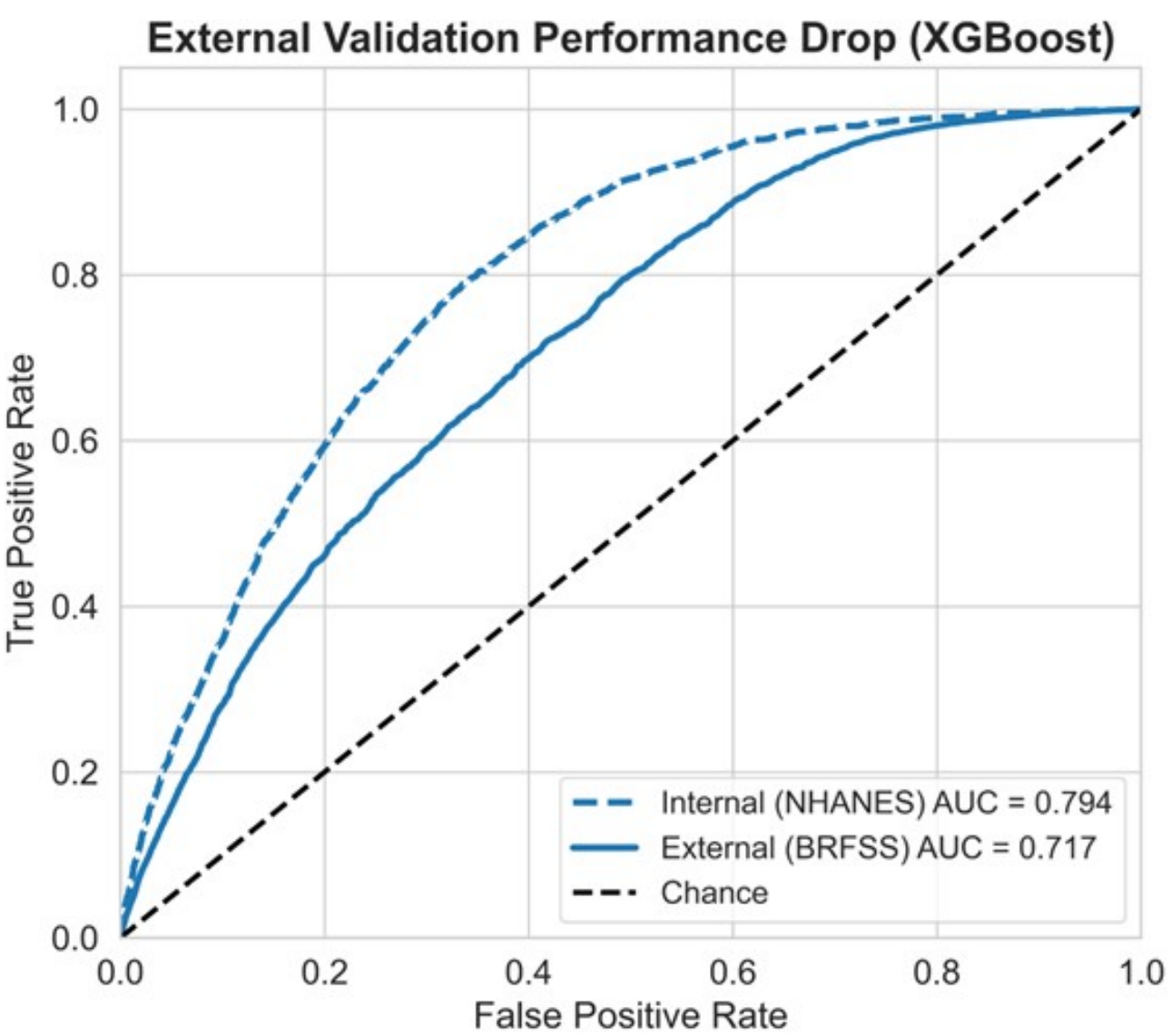

Fig. 2. External validation ROC curve on BRFSS showing performance degradation relative to internal validation.

### D. Model Interpretability

Feature importance ranked by mean absolute SHAP value over all predictions on the external validation cohort was identified using SHAP analysis. The most important predictors were Age (mean |SHAP| = 0.142), BMI (0.098), Physical Activity (0.067), Race/Ethnicity (0.054), History of heart

attack if any (0.046), History of stroke (0.041), gender (0.032), and smoking status (0.029). All directional associations were consistent with accepted clinical evidence, thereby providing model face validity. Older age and higher BMI increased predictions of diabetes risk, while physical activity exerted protective effects. Previous history of heart attack and stroke increased risk, mirroring epidemiologic data associating car- diovascular disease with insulin resistance.

Figure 3 shows the SHAP summary plot, which demonstrates feature contributions to diabetes risk predictions.

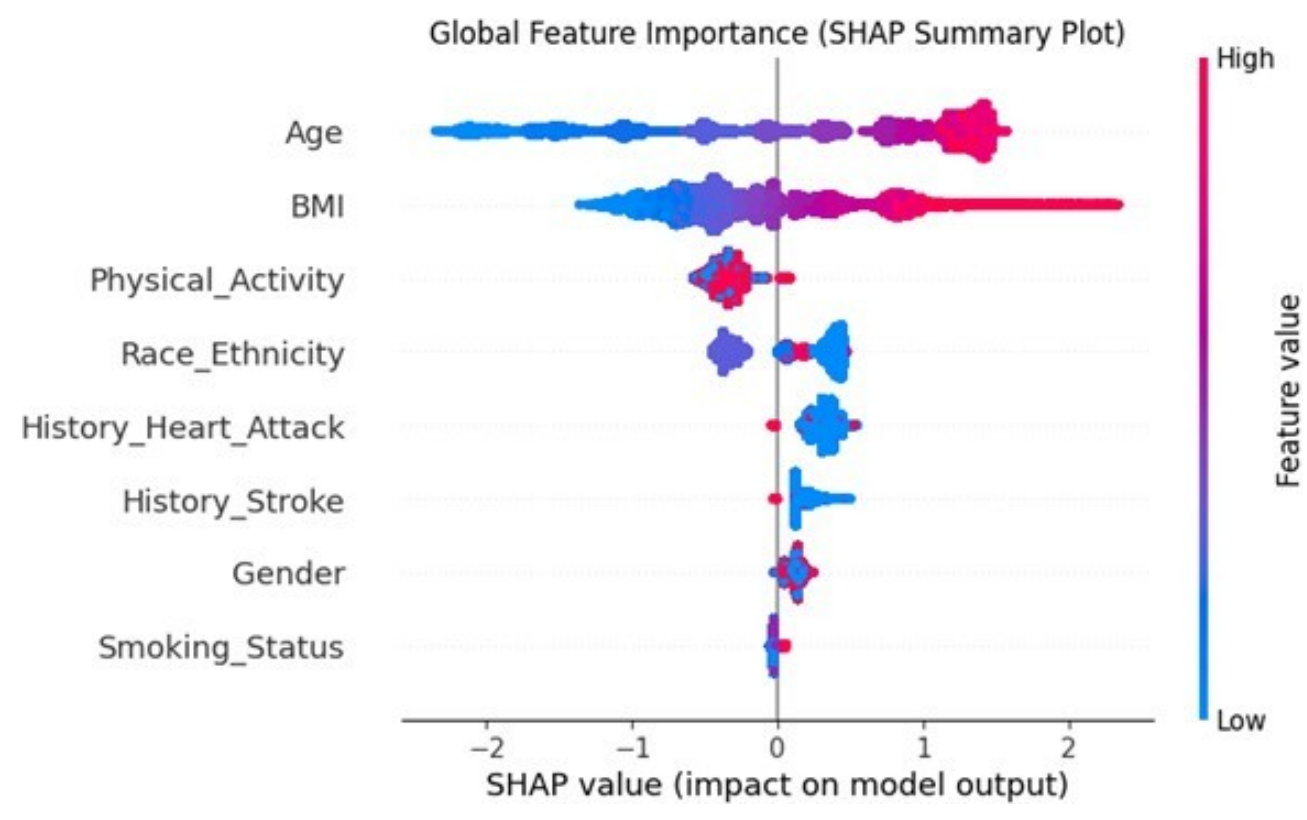

Fig. 3. SHAP global contribution summary plot for XGBoost on the BRFSS external validation data.

### *E. Comprehensive Fairness Analysis*

Table III presents fairness evaluation of the demographic and clinical subgroups, which showed significant performance disparities that can be masked by combined metrics.

TABLE III
FAIRNESS ANALYSIS ON EXTERNAL VALIDATION (BRFSS)

| Subgroup | Category | n | AUC (95% CI) | Δ vs Ref | p-value |
|---|---|---|---|---|---|
| **Overall** | — | 1,285,783 | 0.717 (0.712–0.722) | — | — |
| **Gender** | Male | 595,462 | 0.723 (0.721–0.725) | Reference | — |
| | Female | 687,435 | 0.712 (0.710–0.714) | 0.011 | 0.142 |
| **Age Group** | 18–39 yrs | 285,445 | 0.742 (0.738–0.746) | Reference | — |
| | 40–59 yrs | 512,338 | 0.728 (0.725–0.731) | 0.014 | 0.031 |
| | ≥60 yrs | 488,000 | 0.607 (0.604–0.610) | 0.135 | $< 0.001$ |
| **BMI Category** | Normal (18.5–24.9) | 287,445 | 0.735 (0.731–0.739) | Reference | — |
| | Overweight (25–29.9) | 445,231 | 0.718 (0.715–0.721) | 0.017 | 0.006 |
| | Obese (≥30) | 553,107 | 0.698 (0.695–0.701) | 0.037 | $< 0.001$ |

There was no significant difference in algorithmic classification performance among male (AUC 0.723) and female participants (AUC 0.712) ($p = 0.142$), suggesting comparable performance between genders. Potential reasons for this include that (similar to the observed 48.5% and 51.5% male/female representation in training data), there are few sex differences between features and outcomes.

The youngest adults (18-39 years, AUC 0.742) discriminated the risk level significantly better than elderly adults (≥60 years, AUC 0.607), with a difference of 0.135 AUC ($p < 0.001$). This difference of 13.5 points demonstrates that the performance of the model is hardly better than chance for the elderly, albeit having the highest absolute diabetes risk. This paradox has serious clinical implications: the population at high risk of developing diabetes is the least assessed by the algorithm.

The performances to assess obesity were significantly better for normal weight subjects (AUC 0.735) compared to obese ones (AUC 0.698, difference: 0.037, $p < 0.001$). As with age discrepancies, this dichotomy is alarming, as those people who are obese and thus at greater risk for diabetes receive the worst algorithmic performance.

Figure 4 visualises these fairness disparities through forest plot representation, highlighting the severe degradation for vulnerable subgroups.

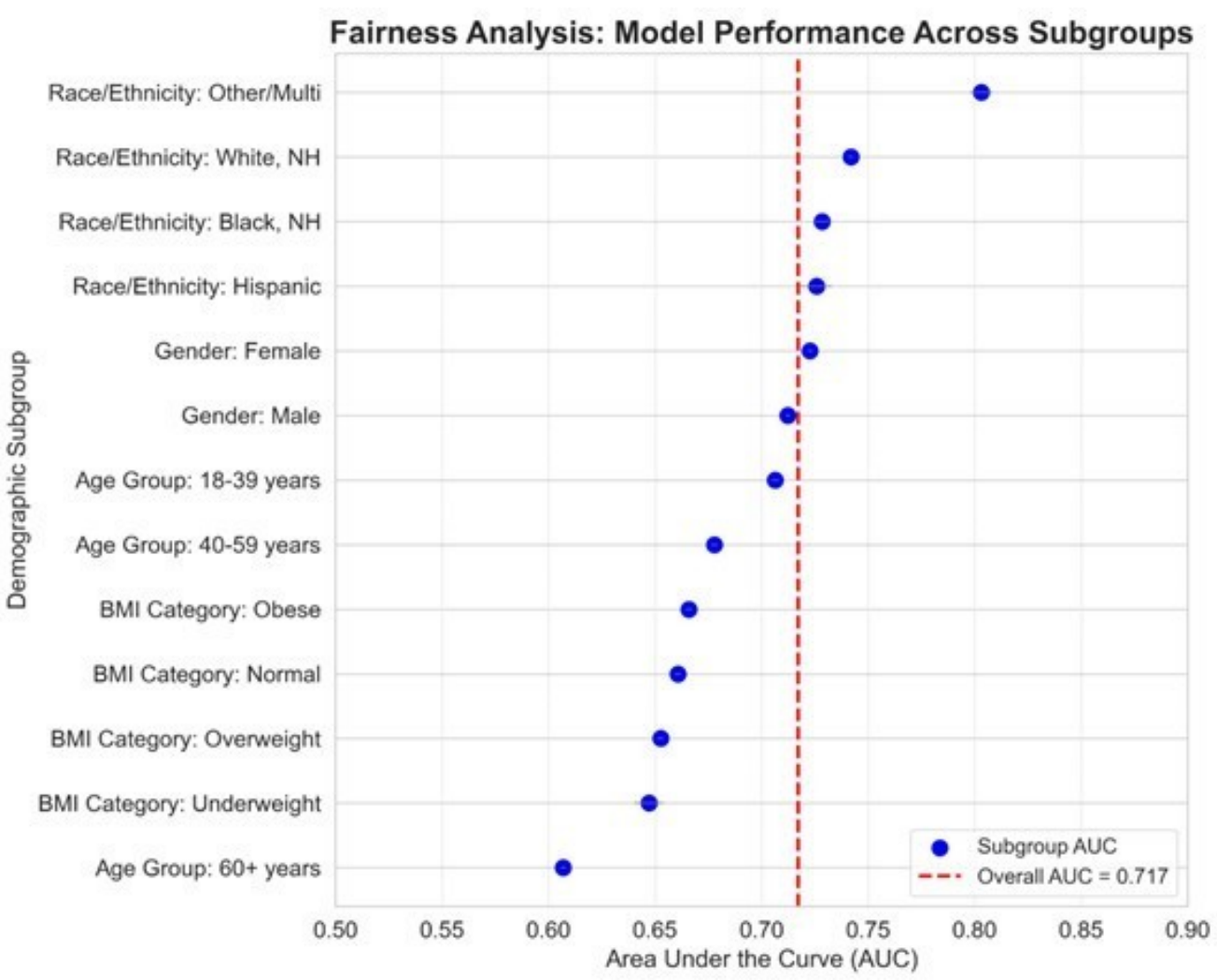

Fig. 4. Fairness analysis showing AUC performance across demographic and BMI subgroups on BRFSS.

### *F. Calibration and Clinical Utility*

Calibration analysis disclosed all-encompassing miscalibration on external validation. A moderate degree of the calibration error was diagnosed with a Brier score of 0.123 (95% 0.118–0.128). Decile calibration curves indicated an exaggeration of risk at high thresholds: individuals in the highest decile (predicted probability $> 0.5$) had observed diabetes frequencies of 0.40-0.45, suggesting overestimation by 10-15 percentage points. This mis-calibration trend may indicate that the model learned risk associations particular to the NHANES population did not generalise completely to BRFSS. Clinical use would necessitate recalibration with isotonic regression or Platt scaling to re-calibrate predicted probabilities and align model sensitivity specifically for the study population due to observed frequency [13].

Performing decision curve analysis, the model generated net benefit across risk thresholds from 10% to ∼18% (consistent with typical risk thresholds for diabetes screening [5]). With this range, the model-guided screening is better than universal screening and no screening. However, the net benefit turned
negative above thresholds of 18% because a systematic overestimation generates more false alarms than true detections.

## IV. Discussion

### A. Principal Findings and Clinical Implications

This holistic assessment exposes a deep-seated rift between internal validation effectiveness and real-world applicability. Despite high internal discrimination (AUC 0.794), XGBoost was found to deteriorate upon external validation (AUC 0.717, relative deterioration of 9.7%). This level is consistent with systematic reviews reporting standard external drops of 10- 15% for clinical prediction models.

Yet a fairness analysis found more worrying trends. Older adults (aged ≥60 years, AUC 0.607) and obese individuals (BMI ≥30 kg/m², AUC 0.698) had very poor performance despite being populations at the highest absolute risk for diabetes. These differences—corresponding to a 13.5-point and 3.7-point AUC gap, respectively—would reinforce patterns of health inequality if implemented without intervention.

Indeed, with a Brier score of 0.123 and systematic overestimation at high-risk scores, predicted probabilities are not in a form that can be considered as usable estimates of true risks without adjustment to the average future risk in the population of interest ('recalibration'). The prediction of e.g., a diabetes probability of 60% could end up matching an observed frequency in data as low as 45%, leading to wrong clinical decisions, undue anxiety for the patient and ineffective allocation of resources.

The most alarming discovery is that populations with the highest absolute diabetes risk experience the worst algorithmic evaluation. Diabetes prevalence nears 25-30% in the elderly, but the model performs poorly for this high-risk group (AUC 0.607). This is likely due to several factors: age-associated feature interactions not captured in training, complexity of co- morbidity patterns in elderly populations generating noise, and increased measurement error in self-reported health conditions among older adults.

A few potential interventions also deserve further investigation: a new model that is trained in separated younger and older groups, a fairness-aware training method with added constraints of demographic parity or equalised odds, through oversampling elderly and obese people so the subgroup imbalance could be less during the training process, and post-hoc calibration by vulnerable subgroups [16].

### B. Contextualization with Existing Literature

Our results broaden recent extensive validation studies, offering new insights. Deberneh and Kim reported close AUC degradation in an external setting (from internal AUROC=0.85 to external AUROC=0.73) on a dataset of 253,000, but did not perform the fairness analysis [14]. Vastly larger external validation (n=1.28M), systematic fairness auditing with dis- covered age and BMI disparities, multi-dimensional assess- ment incorporating calibration and interpretability, transparent TRIPOD-AI methodology contributing to reproducibility, are only possible due to the ongoing evolution of methods and technology in this space, completing these necessary early steps.

Lai et al. reached an external AUC 0.74 with electronic health records, which is in the same range as our 0.717 [15]. Nevertheless, their reliance on laboratory-heavy EHR data restricts generalizability to the screening setting. Our in-field approach allows deployment to the community, pharmacies and telemedicine settings where blood testing is not available.

The age-related fairness discrepancy we observed mirrors observations in other clinical ML domains. Observation of algorithmic failures in predictive model studies, such as for the prediction of sepsis and surgical risk assessment, has also shown for older adults further indicates that age-bias deficiencies are systemic across fields.

### C. Limitations

Several limitations warrant acknowledgement. First, the BRFSS race/ethnicity had 44.2% missingness, which was sufficient to preclude full equity analysis for this important demographic dimension. With demonstrated racial/ethnic disparities in diabetes prevalence and access to care, this is a sizable gap that merits further exploration with more complete data.

Second, both cohorts include the same periods (2015-2022), which means there are limitations on how we can assess temporal validity. As time horizons extend, population attributes, clinical practices, and diagnostic criteria change, which may lead to diminishing returns in model performance. Gaps in temporal validation of 5-10 years would be necessary to justify greater confidence in affirming longitudinal validity.

Third, the restriction to non-laboratory predictors of necessity means that the discrimination would be lower, relative to models that did include some form of biomarkers. Models that are published, which include HbA1c and fasting glucose, are known to achieve AUCs greater than 0.90, though a blood draw would be needed, which is not feasible in many of the screening scenarios. Accessibility was paramount in deciding this tradeoff.

Fourth, BRFSS is reliant on respondents' self-reported diabetes diagnosis, which presents the possibility of misclassification bias. Undiagnosed cases are completely omitted, and the accuracy of self-report varies across demographic groups, including education, health literacy, and access to healthcare. This measurement heterogeneity does exist and would not be something unusual to the development and validation cohorts, but it does make the challenges of deploying the models more intricate.

Fifth, we did not prospectively apply fairness-aware training methods. Our fairness analysis was performed retroactively, after the inequities were identified. Incorporating fairness constraints at the model development stage in future works will prevent inequities from occurring in the first place.

### D. Future Work and Deployment Recommendations

Multi-site external validation comprising various healthcare systems, including safety-net hospitals and rural clinics, and internationally, will be valuable in future work. The geographic and socioeconomic diversity will provide further evidence on the performance heterogeneity, which has not been captured by the NHANES-BRFSS validation.

We will be able to ascertain the clinical utility of the model through prospective validation with randomised trials, where model-guided screening will be compared to standard care. Such trials should focus on new downstream outcomes, such as new diabetes diagnoses, time to diagnosis, and uptake of preventive interventions, and not on the surrogate discrimination metrics.

For the identified fairness gaps, an age and BMI-stratified model could be of interest. There may be an improvement in the within-group performance in models trained separately for different demographic groups, though it could be more difficult to deploy. Fairness-aware methods of training may help reduce these gaps while still allowing a unified model of the deployment.

The minimum standards we advocate for before any clinical deployment include rigorous external validation demonstrating sustained AUC $> 0.70$, calibration evaluation with recal- ibration for Brier score greater than 0.15, fairness being assessed with complete transparency, communication with other medical professionals that these forecasting tools are screening tools that aid triage and are not final verdicts, monitoring continuum with sustained quarterly evaluations and annual recalibration, and TRIPOD-AI compliance for outcome reporting.

## V. Conclusion

This large-scale evaluation validated that XGBoost achieved high internal discrimination (AUC 0.794) but experienced performance degradation on external validation (AUC 0.717, representing a 9.7% relative decline). More concerning than this expected performance drop were the severe failures ob- served in vulnerable groups.

The fairness analysis showed that the paradox from these groups was ethically troubling. Older adults aged 60 and above showed an AUC of only 0.607, while obese individuals (BMI $\geq$30 kg/m²) showed an AUC of 0.698. The deployment of these algorithms appears to come with a degree of risk that would disproportionately impact these vulnerable groups. These groups comprise the AUC gaps of 13.5 points and 3.7 points. These scenarios, with the lack of mitigation, would worsen inequities in providing health care.

For developing clinical AI, we suggest four key components: performing large-scale external real-world validation; assessing calibration by measuring and comparing predicted versus observed probabilities; conducting detailed fairness auditing with transparent reporting of subgroup disparities; and implementing comprehensive evaluation beyond discrimination metrics alone.

In any clinical domain that relies solely on internal validation or limited external validation, this is where we draw the line. The critical lesson is to synthesize multiple metrics comprehensively rather than relying on a single metric when making deployment decisions. It is internal validation that lends false optimism when faced with a distribution shift or measurement heterogeneity, plus there is demographic diversity.

In the case of diabetes screening, this research ultimately shows that non-laboratory ML models can deliver acceptable external performance (AUC 0.717) for appropriate deployment in population-level screening. However, the identified age-related and BMI-related inequities underscore the urgent need for fairness-aware design or age-stratified deployment strategies for at-risk populations.

The way forward is from proving it works in one data set to proving it works in all. When deploying clinical AI in real-world healthcare, we must demonstrate that it benefits those who need it most, not just the populations it was trained on. If not, we risk creating and exacerbating health inequities through algorithmic deployment.

## Acknowledgment

The authors thank the Centers for Disease Control and Prevention for their continued presence in keeping NHANES and BRFSS databases accessible to the public. We recognize the participants who provided data for these surveillance systems.